\pdfoutput=1

\documentclass[10pt, a4paper]{article}

\usepackage[final]{lrec-coling2024}
\usepackage{floatrow}
\usepackage{booktabs}
\usepackage{multirow} 
\usepackage{CJKutf8}
\usepackage{amsmath}

\title{Enhance Robustness of Language Models Against Variation Attack through Graph Integration}

\name{Zi Xiong\textsuperscript{1,2,†}, 
Lizhi Qing\textsuperscript{1}, Yangyang Kang\textsuperscript{1,*}, Jiawei Liu\textsuperscript{2,*}, \\ \textbf{\large Hongsong Li\textsuperscript{1}, Changlong Sun\textsuperscript{1}, Xiaozhong Liu\textsuperscript{3}, Wei Lu\textsuperscript{2}}}
\address{
         zi.d.xiong@gmail.com, \{yekai.qlz, yangyang.kangyy, hongsong.lhs\}@alibaba-inc.com, \\ \{laujames2017, weilu\} @whu.edu.cn, changlong.scl@taobao.com, Xliu14@wpi.edu
         \\ \textsuperscript{1}Institute for Intelligent Computing, Alibaba Group
         \\ \textsuperscript{2}School of Information Management, Wuhan University
         \\ \textsuperscript{3}Worcester Polytechnic Institute
         }

\abstract{
The widespread use of pre-trained language models (PLMs) in natural language processing (NLP) has greatly improved performance outcomes. However, these models' vulnerability to adversarial attacks (e.g., camouflaged hints from drug dealers), particularly in the Chinese language with its rich character diversity/variation and complex structures, hatches vital apprehension. In this study, we propose a novel method, CHinese vAriatioN Graph Enhancement (CHANGE), to increase the robustness of PLMs against character variation attacks in Chinese content. CHANGE presents a novel approach for incorporating a Chinese character variation graph into the PLMs. Through designing different supplementary tasks utilizing the graph structure, CHANGE essentially enhances PLMs' interpretation of adversarially manipulated text. Experiments conducted in a multitude of NLP tasks show that CHANGE outperforms current language models in combating against adversarial attacks and serves as a valuable contribution to robust language model research. These findings contribute to the groundwork on robust language models and highlight the substantial potential of graph-guided pre-training strategies for real-world applications.
 \\ \newline  }
 
\begin{document}
\begin{CJK}{UTF8}{gbsn}
\maketitleabstract

\renewcommand{\thefootnote}{\fnsymbol{footnote}}
\footnotetext[1]{Corresponding authors}
\footnotetext[2]{This work was done when Zi Xiong worked as an intern at Alibaba}

\section{Introduction}

The field of natural language processing (NLP) has seen remarkable advancements in recent years, with pre-trained language models (PLMs) like BERT \citep{devlin2018bert} being one of the most widely adopted tools for various applications, such as language generation \cite{radford2018improving}, text classification \cite{sun2019fine}, and entity recognition \cite{andrew2007scalable}. PLMs are trained on vast amounts of text data, enabling them to capture patterns and relationships between words and phrases. Yet there is a major limitation of PLMs that their vulnerability to adversarial attacks can lead to the dissemination of false or misleading information \cite{jiang2019detect,liu2022order,chen2024research}. Adversarial attacks refer to malicious modifications made to the input text, which can cause the language model to make incorrect predictions or misbehave \cite{ebrahimi-etal-2018-hotflip}. This has become a growing concern, as the use of these models in real-world applications continues to increase. In particular, the Chinese language presents unique challenges in terms of adversarial attacks due to its rich variety of characters \cite{jiang2019detect}.

\begin{figure}[!ht]
\centering
\includegraphics[scale=0.5]{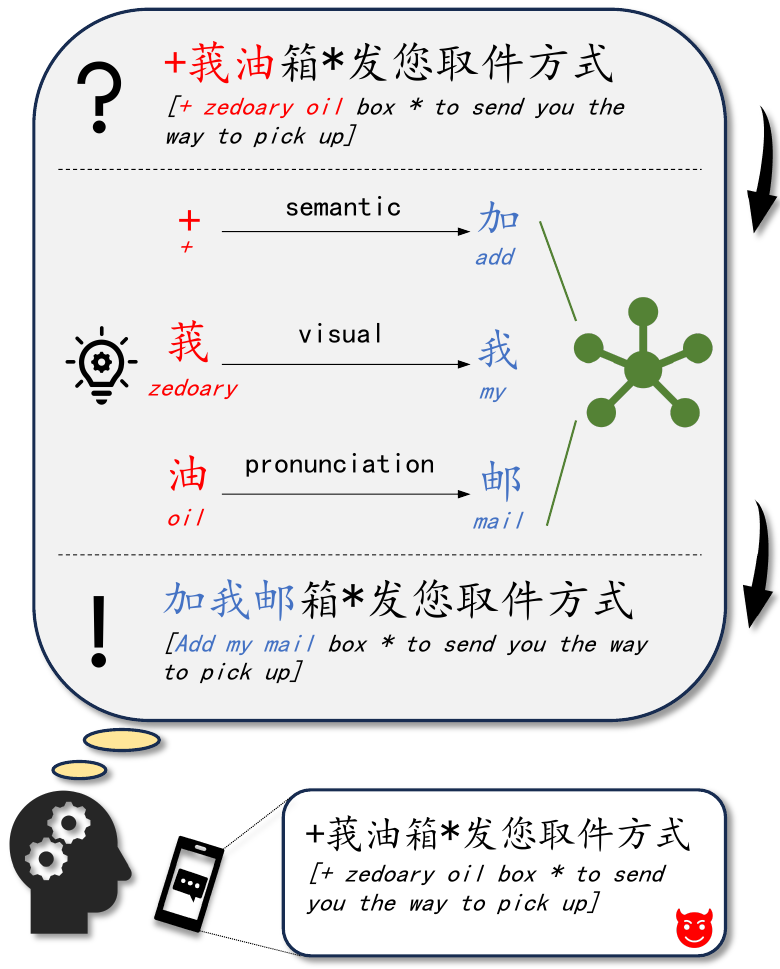}
\caption{Character Variations via semantic, visual, and pronunciation in Chinese Spam Texts.}
\label{fig_intro}
\end{figure}

Existing methods for mitigating the vulnerability of language models to adversarial attacks primarily focus on fine-tuning models with augmented data \cite{wang2020cat}, pre-training models on adversarial examples \cite{su2022rocbert}, employing adversarial training techniques \cite{si2021better} or incorporating regularization methods \cite{liang2021rdrop}. The data augmentation strategy heavily depends on the coverage of the augmented data, which may require an exhaustive exploration of the adversarial space to generate a comprehensive training dataset. This process can be both computationally expensive and time-consuming. Moreover, incorporating adversarial samples may negatively impact the model's performance on clean datasets, as adversarial examples often significantly differ from regular samples.

In this paper, we introduce a novel approach to enhance the robustness of PLMs for the Chinese language. Our proposed method combines multiple adversarial pre-training tasks and incorporates a Chinese Character Variation Knowledge Graph to improve the model's ability to comprehend adversarially attacked natural language text. The multi-task framework facilitates the integration of the knowledge graph into the language model, enabling the model to better capture the linguistic and contextual nuances of the attacked text, thereby enriching the model's textual representations. Our proposed framework, the \textbf{CH}hinese v\textbf{A}riatio\textbf{N} \textbf{G}raph \textbf{E}nhancing method(\textbf{CHANGE}), is illustrated in Figure \ref{fig_framework}. This PLM-independent method bolsters the model's robustness against poisoned text content and consists of two main components:

\begin{figure*}[!t]
\centering
\includegraphics[width=1\textwidth]{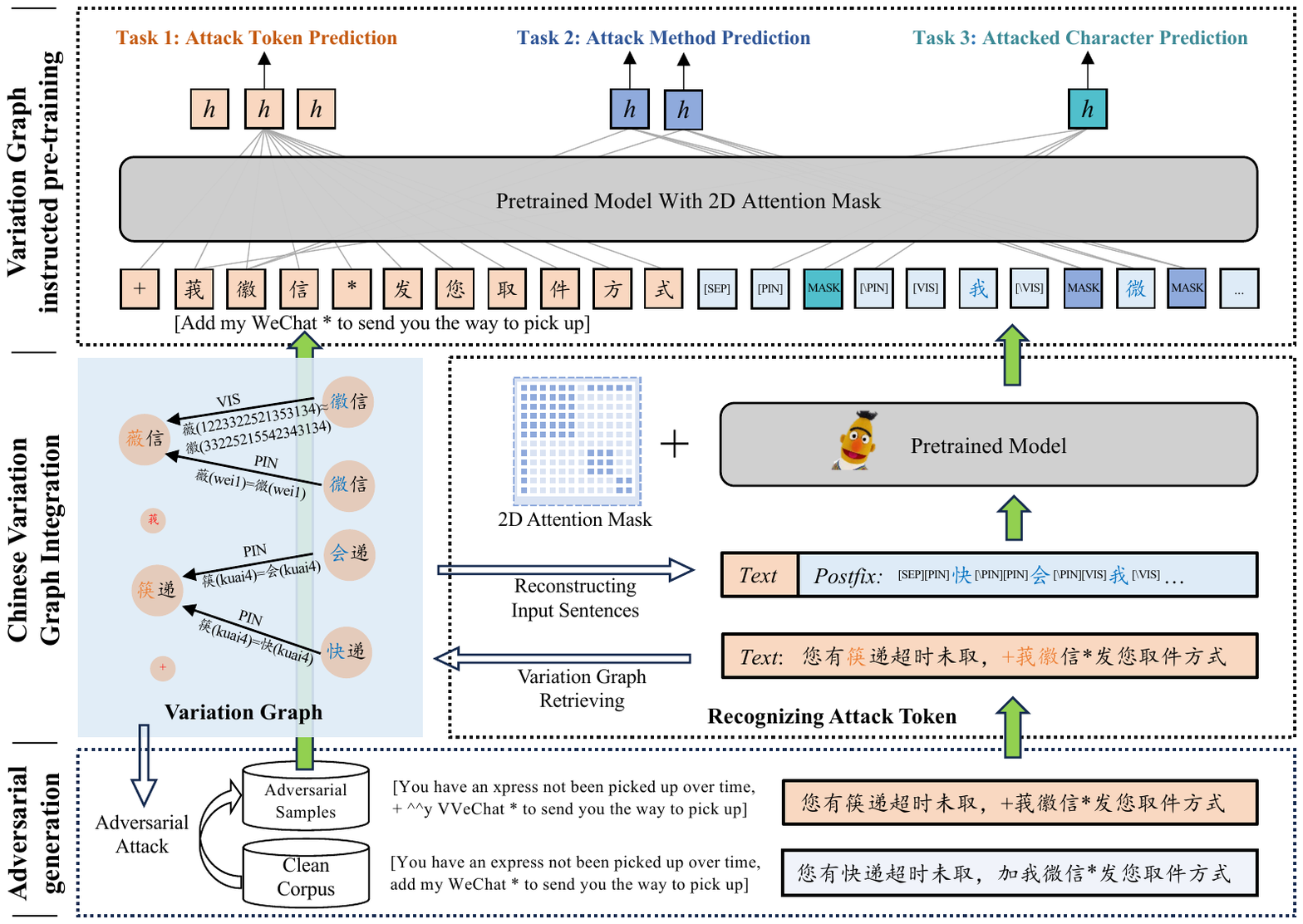}
\caption{The overview architecture of the CHANGE method. For the attacked content, the Chinese Variation Graph Integration recognizes the possible variation and reconstruct a postfix attached to the raw input.}
\label{fig_framework}
\end{figure*}

(1) The \textbf{C}hinese \textbf{V}ariation \textbf{G}raph \textbf{I}ntegration (CVGI) method which employs a variation graph to enhance PLM robustness during the fine-tuning procedure. As depicted in Figure \ref{fig_framework}, we reconstruct the input sentence following the variation graph.

(2) A Variation Graph Instructed Pre-training method which further trains the PLMs under the guidance of the Variation Graph by appending additional pre-training tasks with the graph to transformer-based PLMs, maximizing the potential of the variation graph.

In comparison to existing techniques, our proposed method offers a more lightweight and cost-efficient solution. By leveraging the graph information as a supplement, the approach maintains convenience while minimally impacting the model's performance on clean datasets. It is primarily attributed to the model's reduced reliance on adversarial data and diminished training on perturbed distributions. Meanwhile, the model places greater emphasis on integrating the graph information, leading to a more streamlined and effective learning process.

Our experimental results demonstrate the superiority of the proposed approach compared to existing PLMs. The results show improved performance in several NLP tasks, as well as increased robustness against adversarial attacks. These findings highlight the potential of multi-task and knowledge graph-augmented language models for practical applications and provide valuable insights for the development of robust language models. 
In conclusion, our contribution to the field of NLP research is a novel approach for enhancing the robustness of language models against adversarial attacks in the Chinese language, which can be applied to other languages as well.

\section{Related Work}

\textbf{Robust Chinese Language Models:}
The use of PLMs has revolutionized the field of NLP, allowing for significant improvements in a wide range of downstream tasks without the need for training a new model from scratch. One of the first and most influential of these models is BERT \cite{devlin-etal-2019-bert}, which employs a masked language model objective and next sentence prediction task to learn universal language representations. This approach has been further refined by subsequent models such as RoBERTa \cite{liu2019roberta} and ALBERT \cite{lan2019albert}. In the realm of Chinese NLP, ChinseBERT \cite{cui-etal-2021-chinesebert} has made significant strides by incorporating both glyph and pinyin features of Chinese characters into its pre-training process, achieving state-of-the-art results on many Chinese NLP tasks. Despite these successes, the focus of these models has largely been on improving performance on standard texts, with relatively little attention paid to enhancing their robustness.

In the face of real-world adversarial scenarios, many black-box attacks have been developed under the assumption that the adversary only has query access to the target models without any knowledge of the models themselves \cite{li2018textbugger, ren-etal-2019-generating, garg2020bae}. To defend against these attacks, countermeasures such as adversarial training and adversarial detection have been proposed to reduce the inherent vulnerability of the model. Adversarial training typically involves retraining the target model by incorporating adversarial texts into the original training dataset, which can be seen as a form of data augmentation \cite{si-etal-2021-better, ng2020ssmba}. Adversarial detection involves checking for spelling errors or adversarial perturbations in the input and restoring it to its benign counterpart \cite{zhang2020spelling, bao2021defending}. While these methods have proven effective in the English NLP domain, they are difficult to directly apply to the Chinese domain due to language differences. As a result, many studies have attempted to design specific defenses that take into account the unique properties of Chinese. For example, \citet{wang2018hybrid} and \citet{cheng2020spellgcn} improved Chinese-specific spelling check using phonetic and glyph information. \citet{li2021enhancing} proposed AdvGraph, which uses an undirected graph to model the phonetic and glyph adversarial relationships among Chinese characters and improves the robustness of several traditional models. \citet{su2022rocbert} proposed RoCBERT to enhance robustness by pre-training the model from scratch with adversarial texts covering combinations of various Chinese-specific attacks, which may not be maintained in downstream tasks.

\textbf{Knowledge Graph Enhanced PLMs:}
The enhancement of language understanding in PLMs can be achieved by integrating structured knowledge and linguistic semantics. Recent advancements in Knowledge-Enhanced PLMs (KEPLMs) have uncovered two primary approaches: The first category pertains to the direct incorporation of structured knowledge in PLMs. This category encompasses methods that leverage linguistic semantics inherent in pre-training sentences, and those that utilize entity embeddings derived from structured knowledge. The former technique is exemplified by Lattice-BERT \cite{lai2021lattice}, which pre-trains a Chinese PLM on a word lattice \cite{buckman2018neural}, exploiting multi-granularity inputs to imbue the model with richer semantic understanding. In contrast, the latter technique employs entity embeddings from knowledge encoders, woven into contextual representations to enhance semantic understanding, as exemplified by ERNIE-THU \cite{zhang2019ernie}. The second category comprises methods that involve reformatting and incorporating knowledge information within the PLM framework, either by encoding textual descriptions of knowledge or transforming knowledge relation triplets into text. This category emphasizes the integration of knowledge descriptions and relation triplets into the textual modality. For instance, KEPLER \cite{wang2021kepler} represents the former by jointly encoding pre-training corpora and entity descriptions within a shared semantic space in the same PLM. The latter technique is effectively demonstrated by K-BERT \cite{liu2020kbert} and CoLAKE \cite{sun2020colake}, which transform relation triplets into text and add the transformed text to training samples, circumventing the need for pre-trained embeddings. This paper posits that amalgamating various forms of knowledge information can substantially bolster the context-aware representation abilities of PLMs.

\section{Chinese Variation Graph Integration}

We introduce a variation graph that encompasses a comprehensive collection of Chinese character variations utilized in adversarial attacks on Chinese text. This knowledge graph captures variations in phonetics, character shape, and other aspects of Chinese characters, representing the most prevalent attack methods employed in malicious texts such as those found in black-gray industries and fraudulent activities \citep{jiang2019detect,su2022rocbert,yang2021scalable}. By incorporating this graph into a language model via a carefully re-designed transformer encoder, our approach enhances the model’s resilience to adversarial attacks and preserves its intended functionality. Furthermore, our proposed method is both flexible and lightweight, making it suitable for integration with most transformer-based PLMs.

\subsection{Chinese Character Variation Graph}
The Chinese Character Variation Graph, includes most of the Chinese character variation approaches. The graph is annotated as $\mathcal{G}={(c_0, r_0, c_1), (c_2, r_1, c_3), ..., (c_i, r_m, c_j)}$, in which $c_n$ means attacked character or attack character, $r_m$ means the attack method. The attack methods include the following variation forms:

\textbf{Pinyin}: In the Chinese language, multiple characters may share the same pinyin code representing their pronunciation, making them homophones. Our pinyin variation replaces a Chinese character with one of its homophones. Since a single character may have multiple pronunciations, this can result in homophone variations with different pinyins. We constructed our pinyin variation relation using the pypinyin library\footnote{https://github.com/mozillazg/python-pinyin}.

\textbf{Visual}: Chinese characters are logograms, and their visual appearance conveys a significant amount of information. Visual similarity can sometimes be used to confuse readers about the intended meaning of a text. Our visual transformation is based on the SimilarCharacter dataset\footnote{https://github.com/contr4l/SimilarCharacter}, which calculates Chinese character similarity using the cv2 library\footnote{https://pypi.org/project/opencv-python/}.

\textbf{Character to pinyin}: In Chinese text attack scenarios, replacing a character with its pinyin code is a common tactic used to evade word filtering checks. Our character to pinyin variation is also based on the pypinyin library.

These methods can be used to attack Chinese text, making it difficult for text censorship systems to accurately detect and remove illegal or malicious content while human readers can comprehend the content by association, experience, or metaphor. In the attacking scenario, an attack incident may correspond to a triplet in the variation graph. For example, in Figure \ref{fig_framework}, a node for the character ``微'' (person) could be connected to a node ``薇'', with an edge ``[PIN]'' to indicate the relationship between the character and the method used to alter it, as ``微'' and ``薇'' share the same pinyin pronunciation ``wei1''. In this paper, we refer to this as an attacking path, denoted as $p(n_1, r_1, n_2)$, representing potential paths used to attack text content. The path is used to attack the text to hide the intention of inducing readers to add the attacker's (potentially fraudulent) WeChat account.
Our variation graph can be readily integrated into PLMs using our proposed CVGI method, which we will discuss in the following section.


\subsection{Variation Graph Integration}

In this section, we elaborate on the techniques of our Chinese Variation Graph Integration method, namely \textbf{CVGI}. As presented in Figure \ref{fig_framework}, the method integrates the graph in a transformer-based PLM by reconstructing the input and building a 2D attention mask corresponding to the reconstructed input. 
The model takes a series of tokens, $(x_1, x_2, ..., x_n)$ as input, and use a stacked transformer encoder layer to generate the contextual representations $H_i$. Specifically, we use the following three steps to integrate the graph:

\textbf{Recognizing Attacked Tokens(RAT):} Our approach begins by utilizing language model probability to identify tokens that may have been subject to attack. These tokens serve as the targets for the addition of graph information. Given the context $C=(w_1, w_2, ..., w_n)$, the bert output $f_{w_i}(C)$ and bert vocabulary $V$, the language probability of $w_i$ is:
$$P(w_i | C) = \frac{\exp{(f_{w_i}(C))}}{\sum_{w_j \in V}\exp{(f_{w_j}(C))}}$$
In an input sentence, we can rank the tokens according to their probability and take the lowest $k$\% tokens as possibly potentially attacked tokens $W^a$. This approach is commonly used in Chinese Spelling Correction problems, where language model output probability is a recognized metric for identifying incorrect words or attacked characters within a sentence. Then, we can leverage the Variation Graph to retrieve possible candidate original words $W^c$ for the attacked tokens $W^a=\{w^a_1, w^a_2, ..., w^a_n\}$ in the graph and the corresponding attacking paths $P=\{p_1, p_2, ..., p_n\}$.

\begin{figure*}[!t]
\centering
\includegraphics[width=1\textwidth]{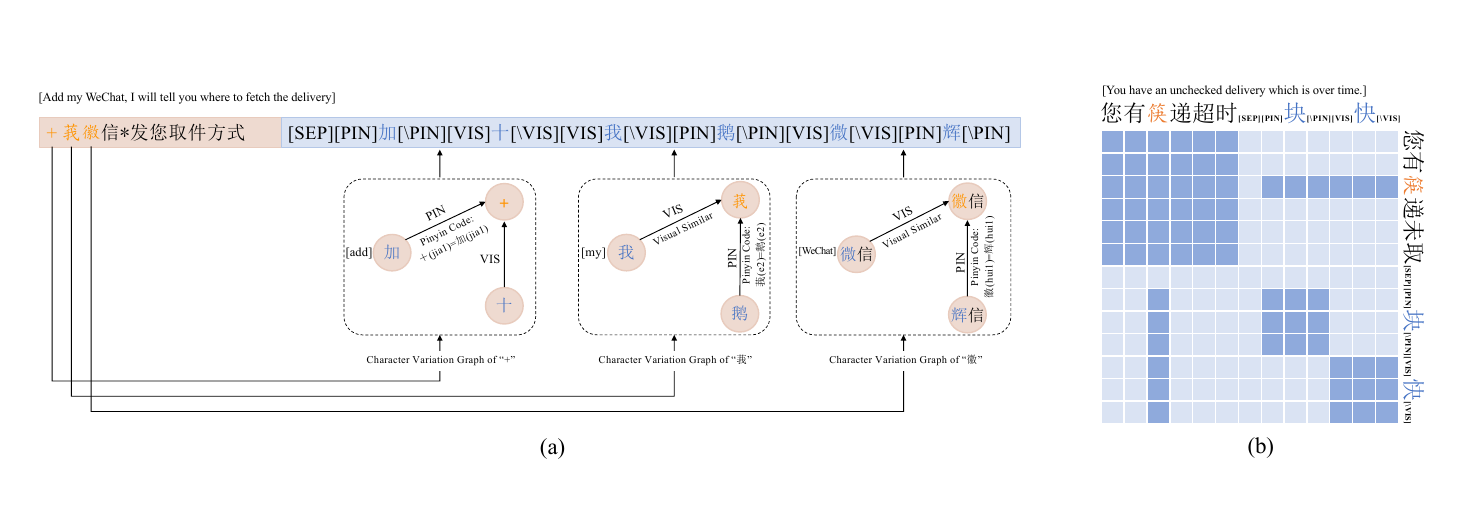}
\caption{(a): An example of reconstruction. In the Variation Graph, the red ``+'' symbol has two variations: ``加'' through pinyin variation and ``十'' through visual variation. Similarly, the red ``莪'' character possesses two variations in the Variation Graph: ``我'' and ``窝'', both derived from pinyin variation. The red ``徽'' character features two variations in the graph: ``薇'' through pinyin variation and ``微'' through visual variation. (b): An example of adversarial 2d attention map. In the whole reconstructed sentence, the raw text segment ``您有筷递超时未取'' employs full cross-attention. The identified attacked character ``筷'' exclusively attends to its variations in the postfix segment. And the candidate original characters, ``[PIN]块[/PIN]'' in example, only have attention with the attacked character ``快''.}
\label{fig_example}
\end{figure*}

\textbf{Reconstructing Input Sentences:} Based on the original input $C$ and the attacking paths $P$ we get, we will add a postfix to $C$ which is generated from $P$. As shown in Figure \ref{fig_example} (a), an attacking path $p($微$, pinyin, $薇$)$ will generate an span formated as ``[PIN]微[/PIN]'' and appended to $C$. Specifically, the [PIN]([/PIN]) token corresponds to \textit{pinyin variation}, while [VIS]([/VIS]) corresponds to \textit{vision variation}, and [CTY]([/CTY]) corresponds to \textit{character to pinyin variations}. Note that each $w^c$ in the possibly attacked tokens $W^c$ may correspond to several attacking paths $p$, which may result in a very long reconstructed postfix and bring much noise to the original input $C$. To reduce the computational complexity and weaken the noise, we further use a 2D mask to eliminate the cross attention between the reconstructed span from attacking paths $p$ and most of the original input $C$, except for only the attack word $w_a$ in $p$.

\textbf{Adversarial 2D attention map:} As shown in Figure \ref{fig_example} (b), our model use a 2D attention mask instead of cross attention mask. For a reconstructed input, the part of the original sentence will use full cross-attention. For the postfix part, ``[VIS]徽[/VIS]'' for example, would only have attention with the attacked token ``微''. The 2D attention mask is calculated during the tokenization phase of the model input. The 2D attention mask is fluently combined with the reconstructed input, enabling the PLMs to gather the original word information to attacked tokens through the attacking path. PLMs have the potential to tell the right attacking path from attention and pick the right original word and inject its information into the original content. 

Our CVGI method has broad applicability to various Chinese NLU tasks and can enhance the performance of most transformer-based PLMs. The reconstruction method is adaptable to various input formats, while the 2D mask is compatible with most of the transformer architectures and the CVGI method does not depend on a specific PLM. The effectiveness of our CVGI method is demonstrated in section 5.3.
Since the objective of fine-tuning does not directly guide the PLMs and the integrated component to construct the attacking path, only fine-tuning is insufficient for the CVGI method to accurately recognize the attacked token and the attacking path from the Variation Graph. We further designed Variation Graph instructed pre-train tasks to help a target PLM to learn to identify the vital path of the graph for enhancing the robustness against attacks in the next section.

\section{Variation Graph Instructed Pre-training}
In this section, we present the details of how we further improve the effectiveness of CVGI by designing Variation Graph instructed pre-training tasks. The key point to strengthen the ability of CVGI is to improve the effectiveness of RAT and the ability to restore the attacking paths by PLMs. It's important to design tasks to help the PLMs learn to adjust the 2D attention mask to the reconstructed text input so that it can better integrate the adversarial information to the attacked tokens. The pre-training is designed to be light weighted so that it can be relatively easy and less costly to apply to ordinary PLMs. In our settings, it only uses a 14 GB corpus from Chinese Wikipedia, Baidu Baike, and THUCTC as the pre-training data.

\subsection{pre-training tasks}
The pre-training task is designed based on the CVGI paradigm to enhance the RAT. We develop a language probability modeling task using Masked Language Modeling (MLM) as the training objectives and create several Variation Graph instructed pre-training tasks. In contrast to common pre-training tasks, our tasks are conducted on attacked samples, which are constructed from a clean corpus using our Variation Graph.
During the attacking process, we generate samples annotated with attacking paths. Next, we reconstruct input examples and produce 2D attention maps similar to CVGI but with slight differences for each task. Specifically, for the RAT enhancing task, we generate an attacked sample with annotations indicating which tokens have been attacked. As we know the ground truth attacking path, we design the reconstructed samples to contain only the true attacking path ($p^T$) and can additionally sample false attacking paths ($p^F$) to construct postfixes and append to the input.
For MLM tasks, we mask tokens in the reconstructed samples following various strategies and conduct MLM training. We design 3 tasks: Attack Token Prediction, Attack Method Prediction, and Attacked Character Prediction, which includes the following:

\textbf{Attack Token Prediction}
Attack Token Prediction (ATP) task is to predict the attack characters based on the raw input, corresponding to the RAT task. Suppose the token $y_i \in {0,1}$ is labeled as attacked (1) or normal (0), its LM probability is $p_i$. The loss of ATP is:
$$\mathcal{L}_{\text{ATP}} = -\frac{1}{N}\sum_{i=1}^{N} y_i\log p_i + (1-y_i)\log(1-p_i)$$
Note that the ATP task can be performed on the target PLM or a certain PLM. It is aimed to enhance the ability of the RAT task by constructing a rich RAT annotated train set using Variation Graph.

\textbf{Attack Method Prediction}
Attack Method Prediction (AMP) predicts the attack method by predicting the masked attack method in the reconstructed input. In the reconstruction, 15\% of the method tokens are masked if there are more than $N$ attacking paths in the reconstructed sample, else one random method will be masked. If the $\hat{y}^M_{i,j}$ stands for the one-hot code of the $i$th method token, $m$ stands for the one-hot code of masked method tokens, $\hat{y}^M_{i,j}$ stands for the probability. The MLM loss of AMP is:
$$\mathcal{L}_{\text{AMP}} = -\sum_{j=1}^{m} y^M_{j}\log(\hat{y}^M_{j})$$
The AMP task aims to predict the relationship based on the head entity and the contextual environment.

\textbf{Attacked Character Prediction}
Attacked Character Prediction (ACP) predicts the attack character by predicting the masked attacked character in the reconstructed input. In the reconstruction, 15\% of the attacked tokens are masked if there are more than 6 attacking paths in the reconstructed sample, else 1 random attacked token will be masked. If the $y^m_{i,j}$ stands for the one-hot for the $i$th attacked character token, $n$ stands for the num of masked attacked character tokens, $\hat{y^c}_{i,j}$ stands for the probability. The MLM loss of ACP is:
$$\mathcal{L}_{\text{ACP}} = -\sum_{j=1}^{n} y^c_{j}\log(\hat{y^c}_{j})$$
The ACP task focuses on predicting the tail entity based on the head entity, the relation, and the contextual environment.

\section{Experiment}
We conducted an evaluation of our proposed method on three distinct datasets and compared its performance against several existing pre-trained language models. Our results demonstrate that our model consistently outperforms the baseline models across all datasets, providing strong evidence for the effectiveness of our approach.
\subsection{Experiment Setup}
\textbf{Training Detail}. 
For pre-training, we utilized a combined corpus comprising Chinese Wikipedia, Baidu Baike, and THUCTC\footnote{http://thuctc.thunlp.org/}. Our model was trained for 100,000 steps with a batch size of 64, a learning rate of 1e-4, and a warm-up rate of 2,000 steps. The corpus contains 13GB of pure text. The training was conducted on 8 Tesla V100 GPUs using the DeepSpeed framework.

\textbf{Baseline Models}.
We tested our approach based on several SOTA models: (1) Chinese-bert-wwm \cite{chinese-bert-wwm}, (2) MacBERT \cite{cui2020macbert}, (3) RocBERT \cite{su2022rocbert}. Chinese-bert-wwm uses the Whole Word Masking pre-training strategy to enhance BERT model especially for the Chinese language. MacBERT applies the MLM-As-Correlation pre-training strategy as well as the sentence-order prediction task. RocBERT is a pre-trained Chinese BERT model that is robust to various forms of adversarial attacks such as word perturbation, synonyms, and typos. It is pre-trained with a contrastive learning objective that maximizes the label consistency under different synthesized adversarial examples.

\floatsetup[table]{font=small}
\begin{table}[!t]
  \centering
  \caption{Experimental results on TNews. Bold shows the best performance of method variants with the same base model.}
  \resizebox{1\textwidth}{!}{
    \begin{tabular}{clcccc}
    \toprule
    \multirow{2}[4]{*}{Base Model} & \multicolumn{1}{c}{\multirow{2}[4]{*}{Method}} & \multicolumn{3}{c}{TNews} & \\
\cmidrule{3-5}          &       & Clean & Att & Argot & \\
    \midrule
    \midrule
    \multirow{3}[2]{*}{\shortstack{Chinese-\\ bert-\\ wwm}}
& BASE & 54.04 & 50.68 & 49.63 & \\
& CVGI & 54.47 & 51.84 & 50.94 & \\
& CHANGE & 54.28 & \textbf{53.91} & \textbf{52.87} & \\
\midrule
\multirow{3}[2]{*}{MacBERT}
& BASE & \textbf{56.12} & 52.34 & 51.54 & \\
& CVGI & 55.33 & 53.49 & 52.62 & \\
& CHANGE & 55.79 & \textbf{55.37} & \textbf{54.22} & \\
\midrule
\multirow{3}[1]{*}{RocBERT}
& BASE & 56.22 & 54.88 & 51.83 & \\
& CVGI & 56.37 & 55.74 & 52.84 & \\
& CHANGE & \textbf{57.09} & \textbf{56.94} & \textbf{54.22} \\
    \bottomrule
    \end{tabular}%
    }
  \label{tab:experiment_tnews}%
\end{table}%

\floatsetup[table]{font=small}
\begin{table}[!t]
  \centering
  \caption{Experimental results on Afqmc. Bold shows the best performance of method variants with the same base model.}
  \resizebox{1\textwidth}{!}{
    \begin{tabular}{clcccc}
    \toprule
    \multirow{2}[4]{*}{Base Model} & \multicolumn{1}{c}{\multirow{2}[4]{*}{Method}} & \multicolumn{3}{c}{Afqmc} & \\
\cmidrule{3-5}          &       & Clean & Att & Argot & \\
    \midrule
    \midrule
    \multirow{3}[2]{*}{\shortstack{Chinese-\\ bert-\\ wwm}}
& BASE & 68.91 & 64.38 & 59.43 & \\
& CVGI & 68.58 & 65.83 & 60.84 & \\
& CHANGE & 69.46 & \textbf{67.96} & \textbf{62.71} & \\
\midrule
\multirow{3}[2]{*}{MacBERT}
& BASE & \textbf{70.83} & 66.75 & 61.50 & \\
& CVGI & 69.98 & 67.67 & 62.40 & \\
& CHANGE & 70.69 & \textbf{68.99} & \textbf{64.20} & \\
\midrule
\multirow{3}[1]{*}{RocBERT}
& BASE & 70.41 & 67.11 & 62.07 & \\
& CVGI & \textbf{70.95} & 67.68 & 62.18 & \\
& CHANGE & 69.85 & \textbf{69.05} & \textbf{63.03} \\
    \bottomrule
    \end{tabular}%
    }
  \label{tab:experiment_afqmc}%
\end{table}%

\floatsetup[table]{font=small}
\begin{table}[!t]
  \centering
  \caption{Experimental results on Message. Bold shows the best performance of method variants with the same base model.}
  \label{tab:experiment_pre-train}%
  \resizebox{1\textwidth}{!}{
    \begin{tabular}{clcccc}
    \toprule
    \multirow{2}[4]{*}{Base Model} & \multicolumn{1}{c}{\multirow{2}[4]{*}{Method}} & \multicolumn{3}{c}{Message} \\
\cmidrule{3-5} & & F1$_\text{micro}$ & Recall & Precision \\
    \midrule
    \midrule
\multirow{3}[2]{*}{\shortstack{Chinese-\\ bert-\\ wwm}}
   & BASE & 82.76 & 91.28 & 75.70 \\
& CVGI & 84.06 & 92.59 & 76.97 \\
& CHANGE & \textbf{85.93} & \textbf{94.66} & \textbf{78.67} \\
\midrule
\multirow{3}[2]{*}{MacBERT}
& BASE & 84.74 & 93.35 & 77.59 \\
& CVGI & 85.69 & 94.33 & 78.50 \\
& CHANGE & \textbf{87.01} & 95.30 & \textbf{80.05} \\
\midrule
\multirow{3}[1]{*}{RocBERT}
& BASE & 85.94 & 95.81 & 77.92 \\
& CVGI & 86.81 & 96.53 & 78.87 \\
& CHANGE & \textbf{87.61} & \textbf{97.37} & \textbf{79.63} \\
\bottomrule
    \end{tabular}%
}
\label{tab:experiment_message}%
\end{table}%

\floatsetup[table]{font=small}
\begin{table*}[!t]
  \centering
  \caption{Ablation experimental results with different base models and different pre-training strategies on TNews, Afqmc and Message. Bold shows the best performance of method variants with the same base model.}
  \resizebox{0.8\textwidth}{!}{
    \begin{tabular}{clcccccccc}
    \toprule
    \multirow{2}[4]{*}{Base Model} & \multicolumn{1}{c}{\multirow{2}[4]{*}{Method}} & \multicolumn{3}{c}{TNews} & \multicolumn{3}{c}{Afqmc} & \multicolumn{1}{c}{\multirow{2}[4]{*}{Message}} \\
\cmidrule{3-8}          &       & Clean & Att & Argot & Clean & Att & Argot &       &\\
    \midrule
    \midrule
    \multirow{4}[2]{*}{\shortstack{Chinese-\\ bert-\\ wwm}}
& CHANGE & 54.28 & \textbf{53.91} & \textbf{52.87} & 69.46 & \textbf{67.96} & \textbf{62.71} & \textbf{85.93} & \\
& \text{ }$\hookrightarrow$w/o ATP & \textbf{54.51} & 52.87 & 51.99 & \textbf{69.61} & 66.99 & 61.88 & 85.07 & \\
& \text{ }$\hookrightarrow$w/o AMP & 53.89 & 52.36 & 51.61 & 68.83 & 66.35 & 61.27 & 84.53 & \\
& \text{ }$\hookrightarrow$w/o ACP & 54.12 & 52.62 & 51.69 & 69.19 & 66.47 & 61.56 & 84.74 & \\
\midrule
\multirow{4}[2]{*}{MacBERT}
& CHANGE & 55.79 & \textbf{55.37} & \textbf{54.22} & 70.69 & \textbf{68.99} & \textbf{64.20} & \textbf{87.01} & \\
& \text{ }$\hookrightarrow$w/o ATP & 55.48 & 54.92 & 53.97 & 70.15 & 68.79 & 63.71 & 86.90 & \\
& \text{ }$\hookrightarrow$w/o AMP & 55.62 & 53.94 & 53.14 & 70.30 & 67.97 & 62.89 & 86.13 & \\
& \text{ }$\hookrightarrow$w/o ACP & 55.68 & 54.19 & 53.29 & 70.58 & 68.19 & 63.08 & 86.27 & \\
\midrule
\multirow{4}[1]{*}{RocBERT}
& CHANGE & \textbf{57.09} & \textbf{56.94} & \textbf{54.22} & 69.85 & \textbf{69.05} & \textbf{63.03} & \textbf{87.61} & \\
& \text{ }$\hookrightarrow$w/o ATP & 56.45 & 55.32 & 53.06 & 69.89 & 67.80 & 62.24 & 86.44 & \\
& \text{ }$\hookrightarrow$w/o AMP & 56.11 & 56.34 & 53.56 & 69.34 & 68.36 & 62.45 & 87.42 & \\
& \text{ }$\hookrightarrow$w/o ACP & 56.44 & 56.43 & 53.27 & 69.72 & 68.25 & 62.64 & 87.29 & \\

    \bottomrule
    \end{tabular}%
    }
  \label{tab:experiment_ablation}%
\end{table*}%

\textbf{Tasks}. 
Our proposed method is tested on two standard Chinese NLU tasks and a real-world toxic detection task, which are: (1) TNews\footnote{https://www.cluebenchmarks.com/tnews\_public}: news title classification with 50k training data, (2) AFQMC\footnote{https://www.cluebenchmarks.com/afqmc\_public}: question matching with 34k training data, (3) Message: The Message Toxic dataset consists of real-world data collected from a real-world application that receives short message notifications and detects messages sent by frauds or black-grey industry practitioners. We manually annotated 15,000 user inputs and identified 2,000 toxic contents (positive), of which 90\% were in adversarial forms. We then randomly sampled 2,000 negative texts and split the entire dataset into training, development, and testing sets with an 8:1:1 ratio.

\textbf{Settings}. 
We conducted the experiment under several circumstances: (1) Clean: the uncontaminated dataset, (2) Att: based on the clean dataset, use the variation graph to attack the data. (3) Argot \cite{ijcai2020p351}: Use the Argot algorithm which is designed for the Chinese language to attack the original data. (4) Toxic: The original toxic detection dataset collected from a real-world application. The test on the contaminated dataset proves the robustness of our method, while the test on the clean datasets proves the generativity of our method. The results are shown in Table \ref{tab:experiment_pre-train}.

\subsection{Experiment Result}
In Tnews, AFQMC and Message datasets, we test the NLU ability of our models under both attacked and clean circumstances, measured by F1-macro. For every task, we report the model performance under 2 adversarial algorithms, Att and Argot. We also report the performance of all base-version models for a fair comparison. The results are presented in Table \ref{tab:experiment_tnews}, Table \ref{tab:experiment_afqmc}, and Table \ref{tab:experiment_message}.
On average, the performance of PLMs using our CHANGE framework negligibly decreased by only 0.05\% on clean data, which is approximately equivalent to no significant change. In contrast, on attacked datasets, the performance of CHANGE improved by 1.21\% (Att) and 1.03\% (Argot), respectively. In the Att and Argot test settings, CHANGE consistently outperforms all baseline models, with particularly notable improvements over MacBERT and Chinese-bert-wwm. 
In the clean dataset, the performance of methods utilizing CHANGE is comparable to that of the baseline models. When employing CHANGE methods, the performance in attacked test sets more closely approximates that in the clean set than when using baseline models. 
Of all the baselines, RocBERT performs closest to RocBERT$_\text{CHANGE}$ under the two attacks, likely due to its multimodal input and its use of adversarial samples during training. Both BERT and MacBERT show significant improvement with CHANGE pre-training and integration. After being enhanced by CHANGE, their robustness performance surpasses that of RocBERT but still falls behind RocBERT$_\text{CHANGE}$. Since BERT and MacBERT were not specifically trained to handle adversarial circumstances, the improvement brought by CHANGE is particularly notable.

\subsection{Ablation Study of CHANGE}

We conducted a series of ablation studies to understand the functionality of different components in CHANGE. In the experiments, The models then tested on three datasets: Tnews, Afqmc and Message. For Tnews and Afqmc, there are three sub-datasets: Clean, Att, and Argot, and the measurement is F1-score. For the message dataset, the measurement is F1-score, precision and recall. All the fine-tuning is conducted with the same base architecture for a maximum of 10 epochs and an early stop strategy on training text.
We devised several variants of CHANGE, each excluding specific components of the strategy: w/o ATP, w/o AMP, and w/o ACP represent variants that exclude the ATP, AMP, and ACP tasks from CHANGE during pre-training.
The results, as illustrated in Table \ref{tab:experiment_ablation}, reveal that the inclusion of ATP, AMP, and ACP tasks led to average performance improvements of 0.53\%, 0.47\%, and 0.61\% for CHANGE, respectively.

\subsection{Robustness Analysis}

As illustrated in Table \ref{tab:experiment_pre-train}, our approach demonstrates effectiveness on both clean and attacked datasets. In clean datasets, our method maintains the original performance of the models, while significantly improving their robustness in attacked datasets. The integration of variation graph-instructed pre-training tasks with our approach results in a noticeable improvement across all attacked experiment settings. In the best-case scenario, the performance improved by 4.3\% in an attacked dataset, reducing the gap between clean and attacked datasets from 5.2\% to 0.9\%. This indicates that our method, CHANGE, can effectively mitigate the effects of attacks on PLMs without sacrificing their performance on normal data.

Moreover, we also conduct an experiment to evaluate the performance of ChatGPT on TNews. Due to space limitations, experiment results are shown in Table \ref{tab:chatgpt}.

\floatsetup[table]{font=small}
\begin{table}[!t]
\resizebox{0.99\textwidth}{!}{
  \centering
  \caption{Experimental results of performance(\%) comparison with ChatGPT on TNews test sets.}
    \label{tab:chatgpt}%
    \begin{tabular}{clccc}
    \toprule
    \multirow{2}[4]{*}{Base Model} & \multicolumn{1}{c}{\multirow{2}[4]{*}{Method}} & \multicolumn{3}{c}{TNews} \\
\cmidrule{3-5}          &       & Clean & Att & Argot\\
    \midrule
    \midrule
\multirow{1}[2]{*}{RocBERT}
& BASE & 55.92 & 54.73 & 51.96 \\
& CVGE & 56.91 & 56.63 & 54.32 \\
ChatGPT & gpt-3.5-turbo-0301 & 43.68 & 40.02 & 37.45 \\

    \bottomrule
    \end{tabular}%
    }

\end{table}%

\subsection{Costs of pre-training}
\begin{figure}[!t]
\centering
\includegraphics[width=1.05\textwidth]{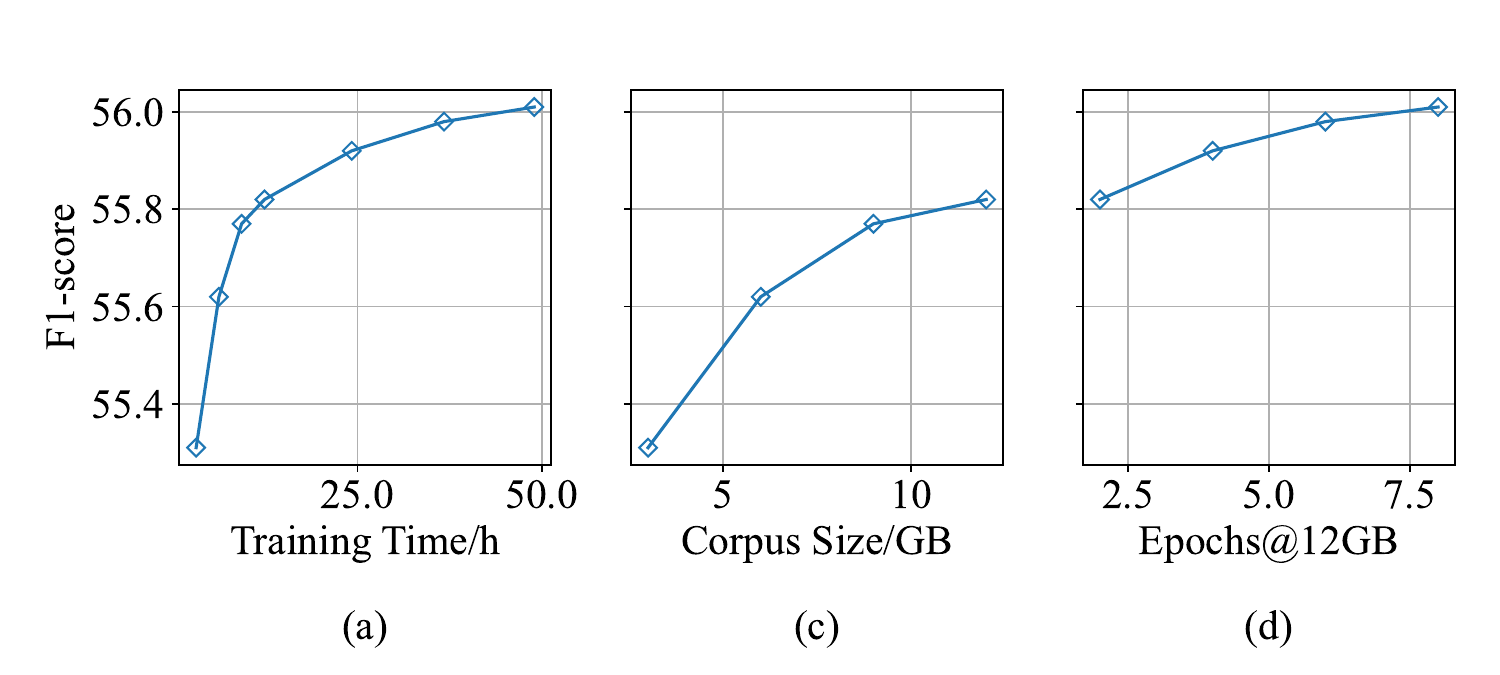}
\caption{The impact of corpus size and training costs on the f1-score performance of CHANGE-enhanced PLM on the TNews Dataset.}
\label{fig_cost}
\end{figure}

Our method follows a plug-and-play design to flexibly and generically enhance the robustness of PLMs. As such, it is essential that our model incurs low training costs. We conducted experiments on the TNews dataset using the Chinese-bert-wwm model. In one set of experiments (Figure \ref{fig_cost} (b)), we trained CHANGE using 3GB, 6GB, 9GB, and 12GB of corpus data (evenly split across multiple sources) for a fixed 2 epochs. In another set of experiments (Figure \ref{fig_cost} (c)), we trained on 12GB of corpus data for 2, 4, 6, and 8 epochs. The training time costs and the effectiveness of fine-tuning are shown in Figure \ref{fig_cost} (a). As can be seen, using 9GB of corpus data yields satisfactory results. Training on 12GB of corpus data for 10 hours on a bert-based pre-train language model is sufficient to achieve strong robustness enhancement. These experiments demonstrate the plug-and-play nature and usability of our CHANGE method.

\section{Limitation}
In this section, we discuss the limitations of our proposed method. Although our approach demonstrates promising results in terms of performance and robustness, several challenges need to be addressed in future research:
(1) Domain Adaptability: While our method exhibits improved performance on various NLP tasks, these tasks may not cover the complete range of domains that pre-trained language models might encounter in real-world applications. Therefore, the adaptability of the proposed approach to the adversarial robustness across different domains warrants further investigation.
(2) Limited Adversarial Defense Scope: Our method enhances the adversarial robustness of pre-trained language models on several NLP tasks; however, potential attack strategies not covered in the experiments might exist. To comprehensively assess the robustness of our approach, it is essential to validate it under a broader range of attack scenarios.
(3) Scalability: Our study focuses on the robustness of pre-trained language models in the Chinese context. However, due to structural and linguistic differences between languages, directly applying the proposed method to other languages may pose challenges. Consequently, appropriate adjustments and validation are required before extending our approach to other languages.
(4) Computational Cost: Our proposed method necessitates constructing the Variation Graph and employing multi-task learning during the pre-training and fine-tuning processes. This might lead to increased computational costs, limiting the applicability of our method in resource-constrained environments.

In conclusion, despite our achievements made in enhancing the robustness of pre-trained language models, several limitations and challenges remain to be addressed. Investigating these issues will contribute to a better understanding and utilization of knowledge graphs and multi-task learning methods, ultimately improving the robustness of language models in practical applications.

\section{Conclusion}

In this paper, we introduce CHANGE, a universal method for integrating the Chinese character variation graph into Chinese language models to enhance their robust representation. Our approach involves designing a method for injecting the graph into transformer-based PLMs during the fine-tuning phase and enhancing adversarial graph injection during PLM pre-training. Experimental results demonstrate that all PLMs enhanced by our CHANGE outperform their respective baselines in robustness tests. Further analysis reveals that CHANGE can effectively capture various paths of common Chinese attacks. For future work, we plan to extend CHANGE to encompass additional types of attacks beyond character variation (e.g., substitution) and apply CHANGE to a broader range of downstream tasks.

\section{Acknowledgements}
This work was supported by National Science and Technology Major Project (2022ZD0116204) and Alibaba Group through Alibaba Innovation Research Program.

\nocite{*}
\section{Bibliographical References}\label{sec:reference}
\bibliographystyle{lrec-coling2024-natbib}
\bibliography{lrec-coling2024-example}

\end{CJK}
\end{document}